\documentclass[journal]{IEEEtran}
\usepackage{amsmath,amsfonts}
\usepackage{algorithm}
\usepackage{algpseudocode}

\algtext*{EndIf} 
\algtext*{EndFor} 
\algrenewcommand\algorithmicindent{1.0em}
\usepackage{array}

\usepackage{textcomp}
\usepackage{stfloats}
\usepackage{url}
\usepackage{verbatim}
\usepackage{graphicx}
\usepackage{cite}
\usepackage{subcaption}
\usepackage{enumitem}
\usepackage{booktabs}
\usepackage{tabularx}
\newcolumntype{C}{>{\centering\arraybackslash}X}
\usepackage{multirow}
\usepackage{makecell}
\usepackage{wrapfig}
\begin{document}

\title{ActiveGlasses: Learning Manipulation with Active Vision from Ego-centric Human Demonstration}

\author{Yanwen Zou$^{12*}$, Chenyang Shi$^{1*}$, Wenye Yu$^{12}$, Han Xue$^{13}$, Jun Lv$^3$, Ye Pan${^1}$, Chuan Wen$^{1\dagger}$, Cewu Lu$^{123\dagger}$
\vspace{0.3em}
$^1$Shanghai Jiao Tong University, $^2$ Shanghai Innovation Institute, $^3$ Noematrix Ltd.
\thanks{$^{*}$ denotes equal contribution.}
\thanks{$^{\dagger}$ denotes corresponding authors.}
}



\maketitle

\begin{abstract}
Large-scale real-world robot data collection is a prerequisite for bringing robots into everyday deployment. However, existing pipelines often rely on specialized handheld devices to bridge the embodiment gap, which not only increases operator burden and limits scalability, but also makes it difficult to capture the naturally coordinated perception-manipulation behaviors of human daily interaction. This challenge calls for a more natural system that can faithfully capture human manipulation and perception behaviors while enabling zero-shot transfer to robotic platforms.
We introduce ActiveGlasses, a system for learning robot manipulation from ego-centric human demonstrations with active vision. A stereo camera mounted on smart glasses serves as the sole perception device for both data collection and policy inference: the operator wears it during bare-hand demonstrations, and the same camera is mounted on a 6-DoF perception arm during deployment to reproduce human active vision. To enable zero-transfer, we extract object trajectories from demonstrations and use an object-centric point-cloud policy to jointly predict manipulation and head movement. Across several challenging tasks involving occlusion and precise interaction, ActiveGlasses achieves zero-shot transfer with active vision, consistently outperforms strong baselines under the same hardware setup, and generalizes across two robot platforms.
\end{abstract}


\section{Introduction}

\IEEEPARstart{T}{he} rapid evolution of data-driven robot learning has cemented large-scale, diverse datasets as the critical driver for achieving generalized robotic manipulation \cite{brohan2022rt, zitkovich2023rt, kim2024openvla, intelligence2025pi05visionlanguageactionmodelopenworld}. Yet, as these models grow in capacity, the field faces a severe ``data hunger" crisis. Robotic data is inextricably bound to the physical world, making its acquisition fundamentally slow, labor-intensive, and expensive. The inefficiency of existing physical data collection pipelines is stark: it has been estimated that accumulating a volume of robotic manipulation data comparable to modern foundational AI datasets would take approximately 100,000 years \cite{goldberg2025good}. This stark efficiency gap underscores a critical necessity in the field: scaling robotic intelligence requires more efficient and scalable frameworks for collecting manipulation data.

Leveraging human demonstrations has emerged as a scalable approach for teaching robots \cite{kareer2025emergence}. However, to truly capture human intelligence, we argue that the data collection process must inherently align with human nature across two fundamental dimensions: \textbf{Manipulation} and \textbf{Perception}.

\begin{figure}[t]
\centering
\includegraphics[width=\linewidth]{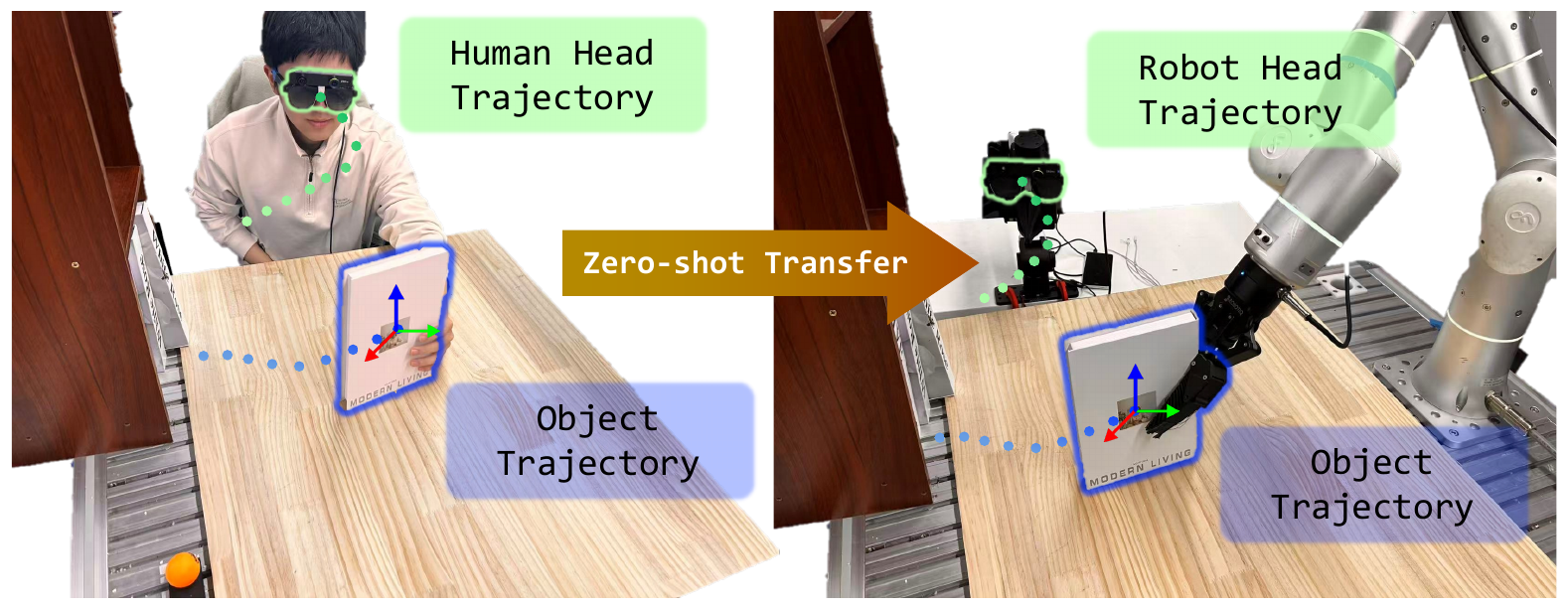}
\caption{\textbf{ActiveGlasses} enables operators to collect manipulation demonstrations with bare hands. The head-mounted glasses record the stereo observations of the current task and the operator’s head movement, and realize zero-shot transfer of manipulation with active vision to robotic platforms. Active vision allows the robot to complete tasks, including occluded scenarios,  with only a head camera input. }
\label{fig:overview}
\end{figure}

From a \textbf{manipulation} perspective, the most natural way for humans to interact with the world is with bare hands. Unfortunately, existing data collection systems typically require operators to teleoperate robot arms~\cite{zhao2023learning,wu2024gello} or rely on bulky handheld devices~\cite{chi2024universal,liu2024fastumi} to perform tasks. While this 1:1 hardware mapping completely bypasses the human-to-robot embodiment gap, it forces the operator to mimic robotic movements, sacrificing our natural kinematic instincts. This compromise is not only physically exhausting (severely hindering scalability), but it also yields constrained, suboptimal data that lacks true human-like smoothness.

From a \textbf{perception} perspective, existing setups suffer from a similar misalignment with human nature. Current paradigms heavily rely on fixed third-person cameras, which are prone to occlusion, or wrist-mounted cameras \cite{cheng2024open,zou2025u}, which serve as a pragmatic hack to provide local visual feedback. However, wrist cameras are inherently passive; their viewpoint is entirely slaved to the end effectors' trajectory. When humans perform complex tasks, we do not rely on passive wrist movements to adjust our perspective. Instead, we possess the ability to actively perceive. We instinctively move our heads to peer around obstacles, lean in for precision, and focus our visual attention independently of our hand movements. By mounting the camera to the robot wrist, prior works discard this rich, intent-driven perceptual signal.

Motivated by the need to capture human-aligned demonstrations, we introduce \textbf{ActiveGlasses}, a lightweight, head-mounted system for robot learning. ActiveGlasses allows operators to collect data entirely in the wild using their bare hands, maximizing comfort, natural kinematics, and scalability. Concurrently, by leveraging the SLAM localization features of commercial AR glasses, the system captures the operator’s 6-DoF(Degree of Freedom) head trajectory. This enables us to record true ``active vision", providing the policy with explicit cues about visual intent and attention.

However, prioritizing natural human behavior during data collection introduces a severe challenge during deployment: the morphological gap. Direct retargeting from a five-fingered human hand and a moving human head to a robotic system is highly error-prone. To bridge this gap, we propose an object-centric, 3D point-cloud policy. Instead of modeling the human’s action kinematics, our policy predicts the 6-DoF trajectory of the manipulated object in the task space, inherently endowing the system with cross-embodiment capabilities. During inference, we utilize a dual-arm setup: a primary arm executes the predicted object trajectory, while a secondary 6-DoF tabletop arm dynamically mimics the human operator's head movements to achieve active perception, overcoming visual occlusions and the difficulty of discerning small objects at a distance.

We evaluate our system on three challenging real-world tasks: book placement, bread insertion, and occluded distant water pouring, which involve significant occlusion and require high manipulation precision. Under the same setting where only a single active perception camera is used as visual input, our method outperforms the baseline $\pi_{0.5}$\cite{intelligence2025pi05visionlanguageactionmodelopenworld} in final success rate by 35\%, 25\%, and 30\%, respectively.

In summary, our system features several innovative designs for scalable data collection and policy deployment:

\begin{enumerate}
\item \textbf{Scalable Bare-Hand Data Collection.} ActiveGlasses employs a commercial AR Glasses setup combined with an on-device GUI for gesture and audio feedback. By eliminating the need for handheld or teleoperation devices, our system untethers the operator, significantly reducing physical burden compared to existing systems.
\item \textbf{Active-Vision-Only Visual Input.} Our framework relies on a single active vision camera throughout both training and deployment. During data collection, it naturally captures first-person human demonstrations. During inference, the system is mounted on a 6-DoF perception arm mimicking human head movement, enabling dynamic handling of occlusions and distant observations without requiring fixed external or wrist-mounted cameras.
\item \textbf{Object-Centric Representation for Cross-Embodiment.} To mitigate the morphological gap between bare human hands and robotic arms without explicit retargeting, we formulate an object-centric 3D policy. By predicting object trajectories from unified point clouds, our policy achieves zero-shot deployment and can seamlessly transfer across different robotic platforms (e.g., UR5 and Flexiv).
\end{enumerate}

\begin{figure*}[htbp]
    \centering
    \includegraphics[width=1\textwidth]{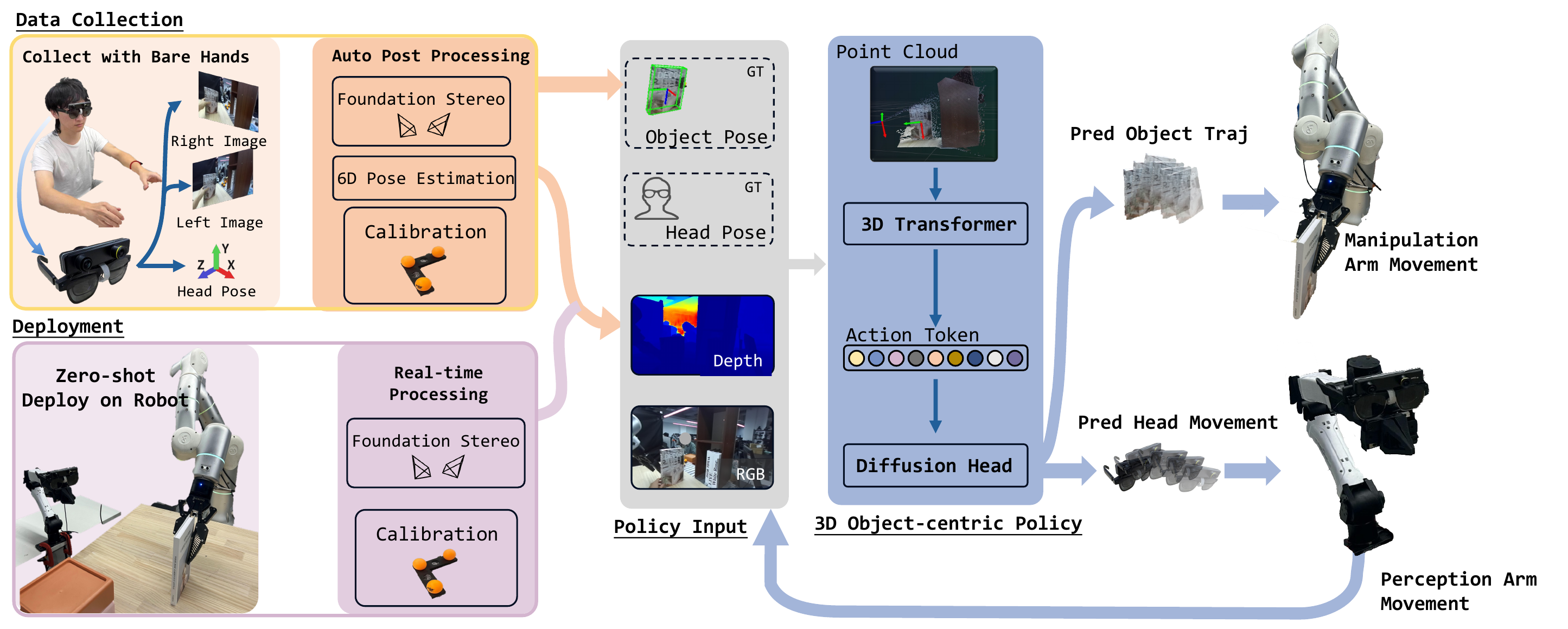}
    \caption{Our system uses XREAL Glasses combined with a ZED Mini stereo camera, enabling egocentric stereo video and 6DoF head movement data collection. During demonstration, the operator actively explores task-relevant regions in the environment and complete manipulation tasks without any hand-held devices. We proposed an object-centric 3D Policy modified from RISE\cite{wang2024rise}, which predicts future 6-DOF object trajectory in the task space. After training, the policy is deployed zero-shot on a real-world robot. A 6-DoF robotic arm synchronously executes the operator’s head motions during inference, allowing the robot to reproduce active vision.}
    \label{fig:system}
\end{figure*}

\section{Related Works}
\textbf{Learning from Demonstration.}
Recent research has increasingly shifted from collecting data via teleoperation\cite{zhao2023learning,wu2024gello,cheng2024open} to leveraging human demonstrations using handheld devices\cite{chi2024universal,liu2024fastumi,wang2024dexcap,fang2025dexop,xu2025dexumi},headset\cite{liu2025egozero,kareer2025egomimic,guzey2025dexterity,yang2025egovla,yuan2025motiontrans,zhu2025emma} or combination of both\cite{chen2025arcap,yu2025egomi} and transferring them to robots. By removing the dependency on the robot platform during data collection, such approaches significantly reduce data acquisition cost and enable in-the-wild data collection.
Handheld devices typically adopt an end-effector configuration and wrist-camera placement similar to those of robotic manipulators. However, these systems rely heavily on wide field-of-view wrist-mounted cameras for perception and therefore lack the ability to perform active sensing in a human-like manner. In addition, such devices often weigh more than 600g\cite{fastumi_pro}, imposing a substantial physical burden on operators during prolonged use. To alleviate the reliance on wrist cameras, other works employ smart glasses\cite{liu2025egozero,zhu2025emma,kareer2025egomimic}, to collect human demonstration data. This, however, introduces a different challenge: the embodiment gap between humans and robots, which is commonly addressed through robot data finetuning\cite{lepert2025masquerade,kareer2025egomimic,zhu2025emma}, action retargetting\cite{yuan2025motiontrans,chen2025arcap,yang2025egovla,zhu2025emma} or visual editing \cite{lepert2025masquerade}, which either limit generalizability or scalability.
Approaches that adopt object-centric representations to align task objectives across different embodiments provide a more general solution\cite{guzey2025dexterity,liu2025egozero}. Nevertheless, in these works, inevitable human head movements are still treated as noise in the policy input rather than as a learnable signal, preventing the model from actively adjusting perception to better accomplish manipulation tasks.

\textbf{Active Vision.}
\cite{bajcsy1988active} argues that active vision is a goal-directed process in which information is acquired through the control of sensing actions. To realize this, some early humanoid designs developed various neck structures\cite{kaneko2008humanoid,kaneko2011humanoid,ishiguro2001robovie,park2005mechanical,park2007mechanical}. With recent advances in learning-based algorithms, some works have explored how to enable robots to achieve active vision through behavior cloning or reinforcement learning, using a camera mounted on a 2-DoF gimbal\cite{cheng2024open,kerr2025eye,xu2026hommi}. However, 2-DoF rotational freedom alone is insufficient for interacting with occluded objects especially in tabletop scenes. Recently, ViA\cite{xiong2025vision} and EgoMi\cite{yu2025egomi} use a 6-DoF robotic arm to address this problem, but the VR headsets they use suffer from limitations in weight, cost, or operator field of view which are constrained by the VR screen.
To the best of our knowledge, \textbf{ActiveGlasses} is the first work to use smart glasses to collect human manipulation data with active vision and achieve zero-shot transfer to robots with an object-centric 3D policy.

\section{The ActiveGlasses System}

Our system aims to enable policies to learn human-like manipulation and active perception behaviors. This requires learning from stable spatial observations, where active vision and manipulation are jointly modeled. To this end, we co-design the hardware and software that allows operators to collect data reflecting natural behavior while preserving rich sensory information. The collected data is further processed into a unified, training-ready representation for policy learning. We introduce these designs in detail as follows.

\subsection{Hardware and Interface Design}
\label{sec:hardware}
When a human performs tabletop manipulation tasks, movements of the upper body as a whole influence the viewpoint of the eyes. Therefore, similar to previous work\cite{yu2025egomi,xiong2025vision}, we use a 6-DoF robotic arm to mimic both neck and torso movements during policy inference. We leverage the 6-DoF pose tracking module of the XREAL Air 2 Ultra to record head motion. Since XREAL does not support direct camera calling due to privacy concern, we adopt a ZED Mini to provide stereo video streams and mount it on the glasses as a substitute.

\begin{figure}[t]
    \centering

    \begin{subfigure}[t]{0.29\textwidth}
        \centering
        \includegraphics[width=\linewidth, height=4cm, keepaspectratio]{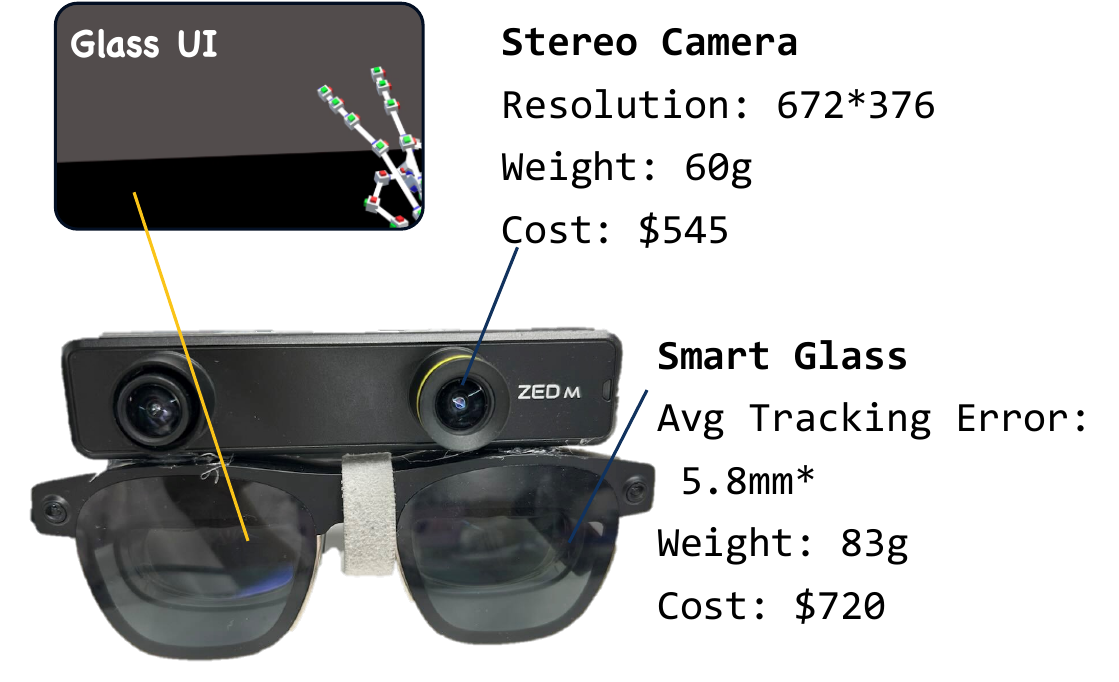}
        \caption{}
    \end{subfigure}
    \hfill
    \begin{subfigure}[t]{0.18\textwidth}
        \centering
        \includegraphics[width=\linewidth, height=4cm, keepaspectratio]{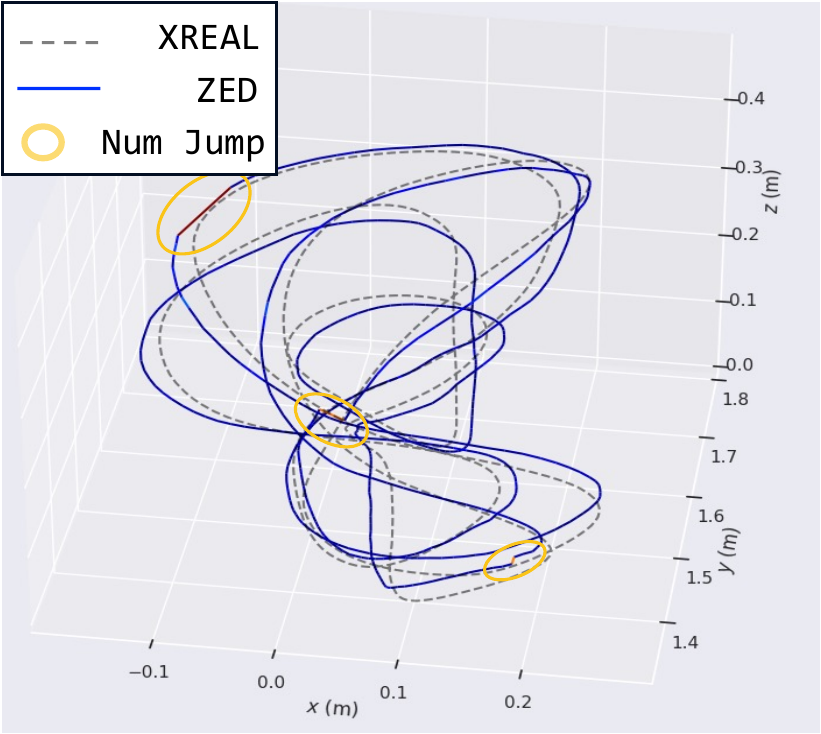}
        \caption{}
    \end{subfigure}

    \caption{
    (a) \textbf{Perception device.} A ZED Mini stereo camera is mounted on smart glasses. No additional external or wrist-mounted cameras are used. Considering the gap between the user's and the ZED camera's field of view (FOV), we added a fixed canvas to the Glass UI to indicate the bottom of the current camera view.
    (b) \textbf{Comparison of recorded trajectories.} Several numerical jumps are observed in the ZED trajectory; therefore, we choose to adopt tracking data from the glasses in training and inference.
    }
    \label{fig:device}

\end{figure}

Note that the ZED Mini also provides an IMU for motion tracking; however, in our tests, the XREAL system demonstrated a higher sampling frequency and greater stability, as shown in Figure~\ref{fig:device}.
    
A user-interface is developed on Unity and shown on glasses. During the data collection process, the glasses will detect the user's gesture as the start\&end signal of one episode with audio feedback. The data recorded in each episode includes the stereo camera frames and the 6-DoF pose from the XREAL Air 2 Ultra. Data timestamps is aligned by ROS.

\subsection{Data Processing}
Our goal is to process the input stereo videos and head trajectory into a unified representation suitable for policy learning. Given the left-eye video 
$V_L=\{l_i\}_{i=0}^{K}$, the right-eye video $V_R=\{r_i\}_{i=0}^{K}$, and the head trajectory 
$H=\{h_i\}_{i=0}^{K}$, the data processing pipeline produces the following outputs in each frame:

\begin{itemize}
    \item Estimation of per-frame depth maps $d_i$
    \item Segmentation of the manipulated object and human hands to obtain masks $m_i^{\text{object}}$ and $m_i^{\text{hand}}$
    \item Ground-truth object trajectory in the left-camera frame $T_{\text{cam},i}^{\text{object}}$
    \item Calibration to estimate transformations from camera and head poses to the world frame, i.e.,
    $T_{\text{world}}^{\text{cam},i}$ and $T_{\text{world}}^{\text{head},i}$
\end{itemize}

\subsubsection{ Depth Estimation and Mask Generation}

Given the stereo image pair $(l_i, r_i)$ at frame $i$, we first estimate the depth map $d_i$ using FoundationStereo\cite{wen2025foundationstereo}. The depth map is then back-projected to reconstruct an RGB point-cloud in the camera frame $p_i^{\text{cam}}$. In subsequent steps, this point cloud will be transformed into a unified world frame.

To remove human-specific visual artifacts, we segment the operator’s hands using Grounded-SAM\cite{ren2024grounded} to obtain the hand mask $m_i^{\text{hand}}$, and remove the corresponding points from the reconstructed point cloud. We also segment the manipulated object using SAM2\cite{ravi2024sam2} to obtain the object mask $m_i^{\text{object}}$, which is used for downstream pose estimation.

\subsubsection{Object Trajectory Estimation}

Object 6-DoF poses serve as the task representation for cross-embodiment deployment. For each frame, we estimate the pose of the manipulated object using FoundationPose\cite{foundationposewen2024}, taking as input the left-view image $V_L$, depth map $d_i$, and object mask $m_i^{\text{object}}$. This produces the object trajectory in the camera frame over the entire episode, denoted as $
\mathcal{O} = \{ T_{\text{cam},i}^{\text{object}} \}_{i=0}^{K}.
$ .

To improve robustness, we also provide the object mesh $\mathcal{M}$ as a geometric prior during pose estimation.

\subsubsection{Calibration}

ActiveGlasses utilizes both the IMU of the smart glasses and the stereo camera inputs. We first perform hand–eye calibration to obtain the fixed transformation between the glasses frame and the camera frame, denoted as $T_{\text{glass}}^{\text{cam}}$. 
However, we observed that the commonly used calibration approach with Aruco markers is prone to be unstable under some head viewpoints, leading to frequent detection failures. We instead introduce a more robust method to establish the world frame, as described below.

We place three orange spheres (as shown in Figure~\ref{fig:system}) on the tabletop that form a planar Cartesian coordinate. These spheres define the origin and the directions of the $x$ and $y$ axes, while the $z$ axis is determined by the right-hand rule.

For a demonstration sequence, we use only the first frame $(l_0, d_0)$ for calibration. The three spheres are segmented using SAM2~\cite{ravi2024sam2}. The 3D positions of the sphere centers in the camera frame are computed from their pixel locations and corresponding depth values, yielding points $b_j \in \mathbb{R}^3, \quad j = 0,1,2.$

We define the world frame using three segmented tabletop spheres with centers
$\mathbf{b}_0,\mathbf{b}_1,\mathbf{b}_2 \in \mathbb{R}^3$ in the initial camera frame.
The axes are constructed as
\begin{equation}
\hat{\mathbf{x}}=\frac{\mathbf{b}_2-\mathbf{b}_1}{\|\mathbf{b}_2-\mathbf{b}_1\|},\ 
\hat{\mathbf{y}}=\frac{\mathbf{b}_0-\mathbf{b}_1}{\|\mathbf{b}_0-\mathbf{b}_1\|},\ 
\hat{\mathbf{z}}=\frac{\hat{\mathbf{x}}\times\hat{\mathbf{y}}}
{\|\hat{\mathbf{x}}\times\hat{\mathbf{y}}\|}.
\end{equation}
Using $\mathbf{b}_1$ as the world origin, the initial camera-to-world transform is
\begin{equation}
T_{\mathrm{cam},0}^{\mathrm{world}}=
\begin{bmatrix}
[\hat{\mathbf{x}}\ \hat{\mathbf{y}}\ \hat{\mathbf{z}}]^\top & \mathbf{b}_1\\
\mathbf{0}^\top & 1
\end{bmatrix}.
\end{equation}

Since the tabletop spheres may be occluded or leave the field of view during the episode and running SAM\cite{ravi2024sam2} for each frame is computational and time consuming, for frame i, we propagate the transform using the head-pose relative motion
$T_{\mathrm{cam},i}^{\mathrm{cam},0}$:
\begin{equation}
T_{\mathrm{cam},i}^{\mathrm{world}}
=
T_{\mathrm{cam},0}^{\mathrm{world}}\;
T_{\mathrm{cam},i}^{\mathrm{cam},0}.
\end{equation}

The entire point cloud is then transformed into the unified world frame. Specifically, an arbitrary point $\mathbf{p}_i^{\mathrm{cam}}$ is mapped as 
\begin{equation}
\mathbf{p}_i^{\mathrm{world}}
=
\left(T_{\mathrm{cam},i}^{\mathrm{world}}\right)^{-1}
\mathbf{p}_i^{\mathrm{cam}}.
\end{equation}

\subsection{Algorithm Design}
We divide a manipulation task into three stages: pre-grasp, motion planning, and termination.

In the pre-grasp stage, we use AnyGrasp\cite{fang2023anygrasp} to perform the grasping action. For tasks that require high-precision grasp poses, a fixed strategy is adopted.

In the motion planning stage, considering ActiveGlasses uses only a single active-vision camera. During demonstration, the operator’s viewpoint varies across episodes since each one starts from a different head pose. This variability introduces significant spatial inconsistency in the 2D image-space observations, while using 3D point cloud as input can maintain consistency. Therefore, following RISE\cite{wang2024rise}, we design a policy that takes point cloud in the world frame as input, and synchronously predicts target object trajectory and headpose movement, as shown in Figure~\ref{fig:policy}. To force the policy to focus on the task objective and avoid the shortcut solution by memorizing the general trajectory, the final policy design does not include the current object pose as an extra condition input to the manipulator diffusion head. We use an absolute representation for object trajectory to align with our 3D policy representation. For the head movement, we adopt a relative representation to avoid the perception arm moving to its workspace limit due to varied initial state of the base, which may further lead to inverse kinematics(IK) failure during policy rollout. Specifically, in the policy, we adopt two diffusion heads to predict the \textit{absolute object trajectory} and the \textit{relative head motion trajectory}, respectively. The detailed ablation study of these choices will be further discussed in \ref{sec:exp}.

Similar to SPOT\cite{hsu2025spot}, we derive the transformation from the object pose to the end-effector pose $
T_{\text{obj}}^{\text{EE}}
=
T_{\text{cam}}^{\text{EE}} \, T_{\text{obj}}^{\text{cam}}
$ through camera calibration to obtain final actions executed by the robot.

Termination is added as an additional dimension to the policy output. In the training dataset, the last five frames of each episode are defined as task completion and assigned a value of 1, while all other frames are assigned 0. The whole pipeline of motion planning and termination is shown in Algorithm~\ref{alg:motion_planning}.

\begin{algorithm}[t]
\caption{Motion Planning with ActiveGlasses}
\label{alg:motion_planning}
\begin{algorithmic}

\State \textbf{Input:} Left camera frames $l_t$, right camera frames $r_t$, horizon $T$

\State \textbf{Output:} Predicted object trajectory 
$\{T_{\text{world}}^{\text{obj}}\}_{t}^{t+T}$, 
headpose trajectory 
$\{T_{\text{world}}^{\text{head}}\}_{t}^{t+T}$, 
termination flag $f_t$

\For{$t = 0$ to $T$}

    \State Estimate depth map $d_t$ from $(l_t, r_t)$
    \State Reconstruct point cloud $p_t^{\text{cam}}$

    \If{$t = 0$}
        \State Calibrate and obtain $T_{\text{world}}^{\text{cam},0}$
    \Else
        \State $T_{\text{world}}^{\text{cam},t}
        =
        T_{\text{world}}^{\text{cam},0}
        \cdot
        T_{\text{cam},0}^{\text{cam},t}$
    \EndIf

    \State Transform to world frame:
    $p_t = T_{\text{world}}^{\text{cam},t} \, p_t^{\text{cam}}$

    \State Clip distant regions of $p_t$

    \State \textbf{Inference:}
    $\big(
    \{T_{\text{world}}^{\text{obj}}\}_{t}^{t+T},
    \{T_{\text{world}}^{\text{head}}\}_{t}^{t+T},
    f_t
    \big)
    = \pi(p_t)$

    \State \textbf{Execute:}
    Robot traj 
    $\left\{
    T_{\text{world}}^{\text{obj}}
    T_{\text{obj}}^{\text{EE}}
    \right\}_{t}^{t+T}$,
    Head traj 
    $\{T_{\text{world}}^{\text{head}}\}_{t}^{t+T}$

    \If{$f_t > \text{threshold}$}
        \State \textbf{break}
    \EndIf

\EndFor

\end{algorithmic}
\end{algorithm}

\section{Experiments}
\label{sec:exp}

We focus on the following questions to evaluate the feasibility of the system:

\begin{enumerate}
    \item \textbf{Scalability.} Compared with other data collection methods (e.g., teleoperation and UMI\cite{chi2024universal}), to what extent can ActiveGlasses improve data collection efficiency?

    \item \textbf{Active Vision.} Compared with a fixed single camera, can active vision effectively improve policy performance? Besides, while improving data collection efficiency and operator comfort, how does ActiveGlasses compare with baseline policies in terms of success rate?

    \item \textbf{Policy Design.} Several design choices in the policy need to be discussed, including whether to use a single or separate diffusion head, whether to include the current pose as an additional condition in the policy, and whether to represent object and head trajectories in absolute or relative form.
    \item \textbf{Cross Embodiment.} Is ActiveGlasses a general data-collection system for various robot platforms? Can we realize zero-shot transfer without masking robots out?
\end{enumerate}

\textbf{Task Design.} We select the following three real-world tasks for evaluation. For each task, we decompose it into three stages (Stage 1–3) to represent task progress:

\begin{itemize}
    \item \textit{Book Placement.} Place a book into a bookshelf. Three books are already placed on the shelf, leaving one empty slot. The camera view is partially occluded by the side wall of the shelf.
    \begin{itemize}
        \item Stage 1: approach the shelf
        \item Stage 2: insert the book into the empty space
        \item Stage 3: complete placement without collision
    \end{itemize}

    \item \textit{Occluded Distant Water Pouring.} Move the teapot to the other side of a screen and position it directly above a cup, then pour water into the cup.
    \begin{itemize}
        \item Stage 1: pass the screen
        \item Stage 2: tilt above the cabinet and align with the cup
        \item Stage 3: pour water into the cup
    \end{itemize}

    \item \textit{Bread Insertion.} Insert a slice of bread into the first slot of a toaster. The slot of the toaster is not visible to the camera at the beginning.
    \begin{itemize}
        \item Stage 1: approach the toaster
        \item Stage 2: move near the slot and align
        \item Stage 3: insert the bread into the slot
    \end{itemize}
\end{itemize}

\begin{figure*}[htbp]
    \centering
    \includegraphics[width=1\textwidth]{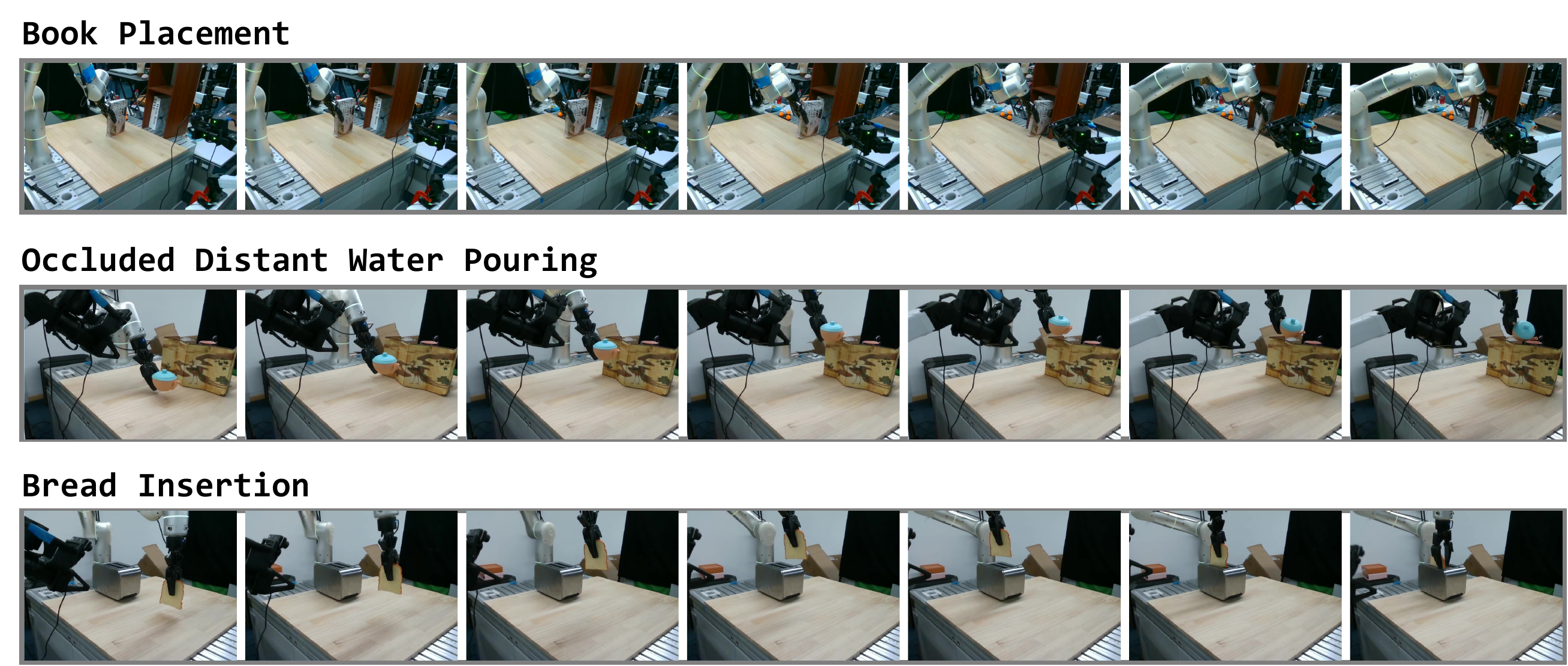}
    \caption{\textbf{Task setting. }We introduce three tabletop manipulation tasks that require active viewpoint adjustment. In \textbf{Book Placement}, the camera initially faces the side of the bookshelf and must move closer and rotate to observe the empty slot before placing the book. In\textbf{ Occluded Distant Pour Water}, the target cup is occluded by a screen, requiring the camera to adjust its viewpoint to perceive the pouring target. \textbf{In Bread Insertion}, the camera must tilt and reorient to observe the toaster slot before accurately inserting the bread. To reflect the operator’s torso movement during data collection, the perception arm is also mounted on a movable wheeled table, and its base position is randomized within a small range at start of each rollout when deployment.}
    \label{fig:tasks}
\end{figure*}

\textbf{Hardware Setup.} We use a Flexiv Rizon4 robot equipped with a Robotiq 2F-85 gripper as the manipulation arm, and an I2RT YAM Robot as the perception arm. A ZED Mini camera together with XREAL Air 2 Ultra smart glasses serves as the perception device mounted on the perception arm via a 3D-printed adapter.

\begin{figure}[htbp]
\centering
\begin{subfigure}{0.45\textwidth}
    \centering
    \includegraphics[width=\linewidth]{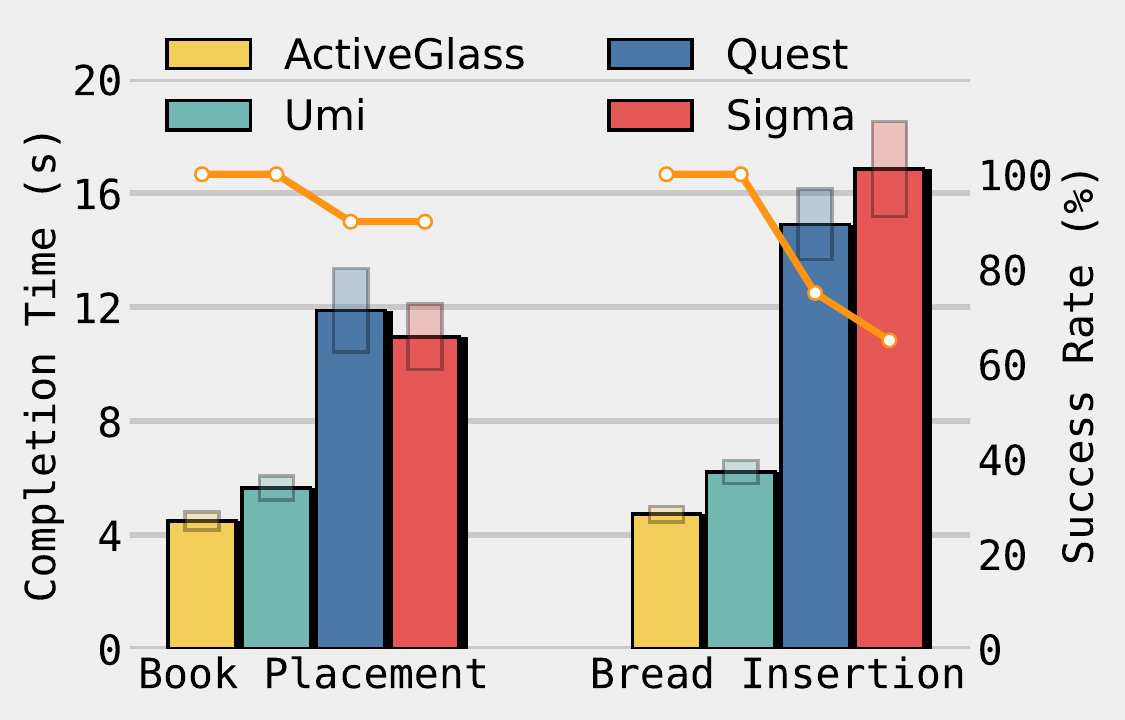}
    \caption{Completion Time (bar) and Success Rate (line)}
\end{subfigure}
\hfill
\begin{subfigure}{0.45\textwidth}
    \centering
    \includegraphics[width=\linewidth]{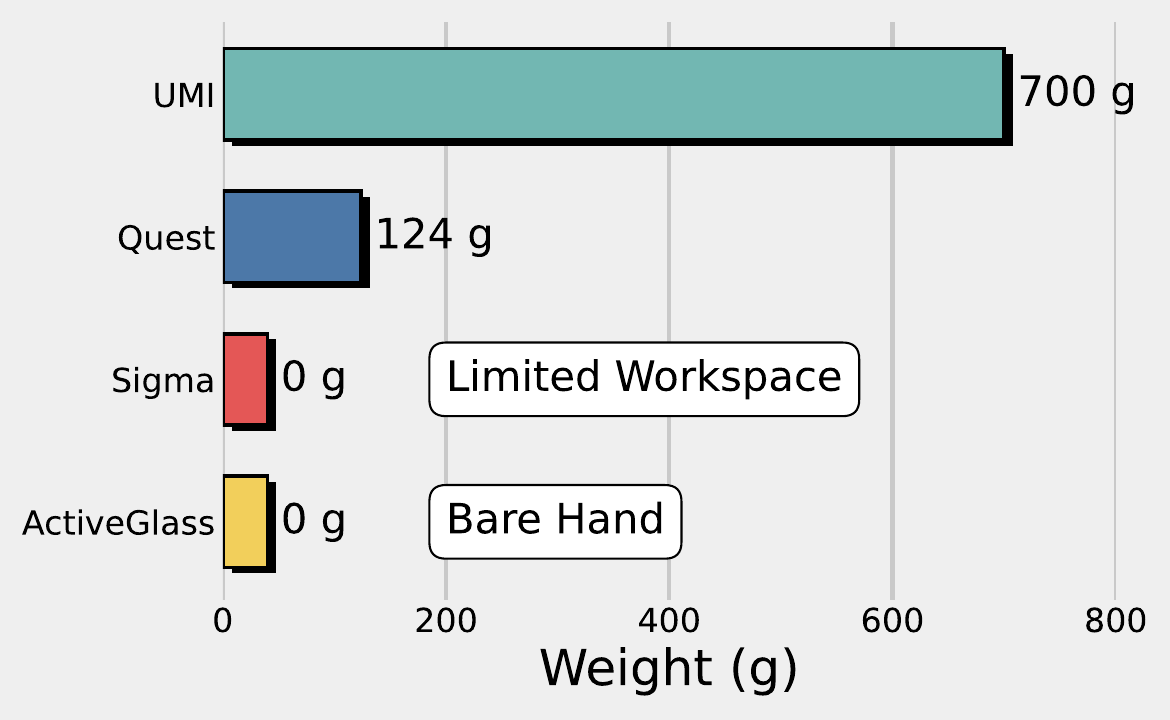}
    \caption{Device Weight}
\end{subfigure}
\caption{Data collection performance. Sigma is reported as 0 g because it operates in a zero-gravity mode.}
\label{fig:data_collection}
\end{figure}

\subsection{Scalability.} We compare ActiveGlasses with UMI\cite{chi2024universal}, VR teleoperation using Quest~3, and the Force Dimension sigma.7 haptic interface in the book placement and bread insertion task in terms of data collection experience. The results are summarized in Figure~\ref{fig:data_collection}.

The gap in visual perception and feedback makes direct human demonstration significantly more efficient and successful than teleoperation. This advantage is particularly evident in \textit{bread insertion}, which requires higher precision. However, the weight of handheld devices can impose a physical burden on operators during prolonged data collection. In contrast, ActiveGlasses combined with bare-hand data collection not only achieves higher data collection efficiency but also provides a more user-friendly experience for the operator.

\subsection{Active Vision}
\begin{table*}[t]
\centering
\small
\setlength{\tabcolsep}{3pt}
\renewcommand{\arraystretch}{1.3}

\newcolumntype{C}{>{\centering\arraybackslash}X}

\begin{tabularx}{\linewidth}{lCCCCCCCCC}
\toprule
 & \multicolumn{3}{c}{\textbf{Book placement}} 
 & \multicolumn{3}{c}{\textbf{Bread insertion}} 
 & \multicolumn{3}{c}{\textbf{Occluded distant water pouring}} \\
\cmidrule(lr){2-4} \cmidrule(lr){5-7} \cmidrule(lr){8-10}

 & Stage 1 & Stage 2 & Stage 3
 & Stage 1 & Stage 2 & Stage 3
 & Stage 1 & Stage 2 & Stage 3 \\
\midrule

\textbf{ActiveGlasses }& 20/20 & 16/20 & \textbf{14/20} & 20/20 & 15/20 & \textbf{11/20} & 20/20 & 15/20 & \textbf{10/20} \\
w/o active vision      & 20/20 & 8/20  & 7/20  & 11/20 & 1/20 & 0/20 & 18/20 & 10/20 & 4/20 \\
Pi05        & 20/20 & 9/20  & 7/20  & 20/20 & 18/20 & 6/20 & 20/20 & 12/20 & 4/20 \\
\bottomrule
\end{tabularx}

\caption{We compare ActiveGlasses with two baselines across three tasks: a variant without active perception and Pi0.5\cite{intelligence2025pi05visionlanguageactionmodelopenworld}. The \textit{Book Placement}, \textit{Bread Insertion}, and \textit{Occluded Distant Water Pouring} tasks are trained using 200, 100, and 100 demonstrations, respectively. During evaluation, the poses of task-related objects are randomized within predefined ranges, including the bookshelf, empty slot location, the cup and screen, and the toaster.
}
\label{tab:main_exp}
\end{table*}
We compare \textbf{ActiveGlasses} with two baselines across 3 tasks, as shown in Tab.~\ref{tab:main_exp}. We remove active vision (\textit{w/o active vision}) while keeping the same policy backbone and action space representation with ActiveGlasses. For Pi0.5 , we train the policy using real-robot demonstrations collected via Sigma teleoperation. During data collection, the operator wears the glasses and moves their head following the task progress to provide corresponding active vision observations. Here the policy is trained under a setting similar to bimanual manipulation. The head-mounted image is used as the only visual input, replacing right wrist camera observation, and left wrist camera input is masked.

The results show that using a single fixed camera leads to poor performance in scenarios involving occlusion and precise manipulation. In the absence of wrist camera observations, Pi0.5 is able to learn certain active vision and manipulation trajectories through proprioception. However, the joint distribution between the manipulation arm actions and the perception arm actions (i.e., the \textbf{head–hand joint distribution}) becomes highly diverse in the 2D image observation space, making it difficult for the policy to extract useful visual patterns in a small dataset. As a result, the manipulation arm tends to ignore the image input and instead execute near-fixed motion trajectories. In contrast, point cloud representation stablizes the visual observation, allowing the policy to effectively model the head–hand joint distribution even with a relatively small dataset and training from scratch. Meanwhile, 6-DoF head movements compensate for the missing wrist camera by purposefully adjusting viewpoints, enabling robust perception and manipulation under occlusion and precise task requirements.

\subsection{Policy Design}
\begin{figure}[t]
\centering
\includegraphics[width=\linewidth]{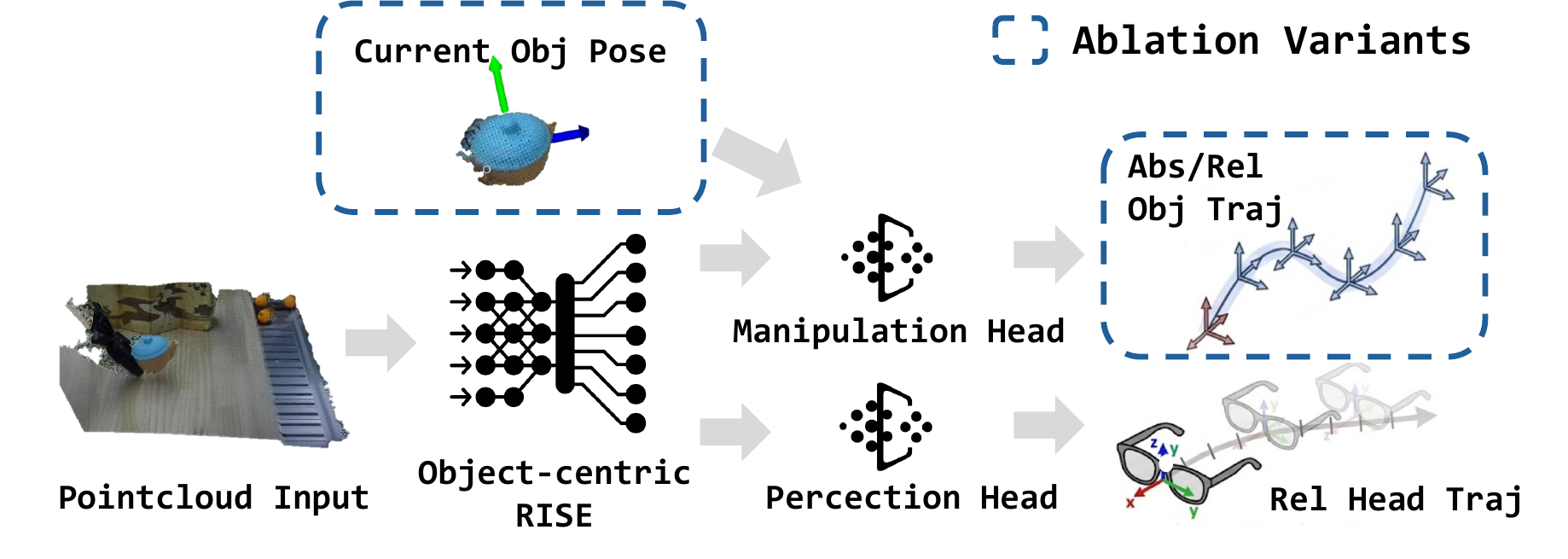}
\caption{\textbf{Policy Design Ablation.} We predict actions of the manipulation arm and the perception head using separate diffusion heads. 
For the manipulation head, we investigate whether the current object pose should be provided as an additional conditioning signal to the diffusion policy. For the predicted object trajectory, we do ablation study on the choice of action representation~(i.e. absolute or relative output).}
\label{fig:policy}
\end{figure}

As shown in Figure~\ref{fig:policy}, we  investigate how the choice of different action representation and extra condition(i.e. current object pose) will influence the policy performance.
Here, we use \textit{abs w/o curr pose} as the default ActiveGlasses setting, and compare it with the following ablations:

\begin{itemize}
    \item \textbf{abs w/ current obj pose}: the manipulation head predicts the absolute object trajectory while additionally conditioning on the object pose of current frame.
    
    \item \textbf{rel w/ current obj pose}: the manipulation head predicts the trajectory with respect to the current frame, while the current object pose is also provided as diffusion condition.
    
    \item \textbf{rel w/o current obj pose}: the manipulation head predicts relative trajectories without explicitly conditioning on the current object pose.
\end{itemize}

\begin{table}[h]
\centering
\begin{tabular}{lcc}
\toprule
 & \multicolumn{2}{c}{\textbf{Book Placement}} \\
\cmidrule(lr){2-3}

 & \makecell{w/o curr pose} & \makecell{w curr pose} \\
\midrule
absolute & \textbf{14/20} & 3/20 \\
relative & -- & 10/20 \\
\bottomrule
\end{tabular}
\caption{We evaluate four policy designs on the \textit{Book Placement} task. An episode is considered successful if the placement is completed without any collision.}
\end{table}

The results imply that for point cloud observations, the correlation between the observation and the action representation is easier for the policy to learn when the trajectory is predicted in absolute representation. In contrast, predicting relative object trajectories leads to a noticeable performance drop. Moreover, the relative representation requires real-time object pose estimation at each step. This not only increases the per-step computing time of the system, but also introduces additional sources of failure. In scenarios where the object pose changes significantly between frames, where severe occlusions occur, or when the object itself is small, it is more prone to losing track of the object, which subsequently causes the policy to fail.

Interestingly, when predicting trajectories in the absolute action space, providing the current object pose as an additional conditioning signal also degrades the policy performance. Instead of relying on the observation to reason about scene changes, it tends to ignore the visual input and learns to execute a nearly fixed object trajectory depending on current object pose. It implies that when outputting absolute object trajectory, explicitly conditioning on the object pose makes it easier for the model to overfit to the dominant motion patterns in the dataset, reducing its reliance on perception. As a result, the generated manipulation trajectory becomes less responsive to variations in the scene.

We also experimented with predicting the absolute head trajectory. However, due to the inherent differences in height and spatial position between the human head and the perception arm's end-effector, the absolute representation often causes the perception arm to move a large distance at the beginning of policy execution. This behavior frequently drives the perception arm closer to the boundary of its workspace, increasing the likelihood of inverse kinematics (IK) failures.

\subsection{Cross Embodiment}

\begin{table}[h]
\centering
\begin{tabularx}{\linewidth}{lCCC}
\toprule
 & \multicolumn{3}{c}{\textbf{Book Placement}} \\
\cmidrule(lr){2-4}

 & Stage 1 & Stage 2 & Stage 3 \\
\midrule

Flexiv Rizon 4 & 20/20 & 16/20 & 14/20 \\
UR5            & 20/20 & 16/20 & 11/20 \\

\bottomrule
\end{tabularx}
\caption{We deploy our policy across two robotic arms and evaluate the policy performance on the \textit{Book Placement} task.}
\label{tab:cross-eb}
\end{table}

The policy predicts the target object 6D pose instead of embodiment-specific representation, and thus inherently ensures cross-embodiment deployment. We evaluate the policy's performance on UR5 in the same setting. The results in Table~\ref{tab:cross-eb} show that the policy achieves comparable performance in first-two stages, while UR5 met more failure cases when trying to place the book. This is because UR5 has a smaller workspace than Flexiv and is therefore less flexible near the limits of its workspace.



\section{Conclusion}

In this work, we propose a novel data collection system, \textbf{ActiveGlasses}, along with a corresponding policy. We mount a stereo camera on smart glasses as the only perception device for both data collection and policy inference. During data collection, the operator simply wears the device and performs tasks with bare hands. Then an object-centric policy takes point clouds as input, and predicts the target object trajectory and the head movement respectively through two separate diffusion heads.

During evaluation, the same hardware is mounted on a 6-DoF robotic arm to mimic human active vision. The system shows zero-shot transfer of manipulation with active vision on three challenging real-world tasks involving occlusion and high-precision manipulation.  Under the same hardware setup, ActiveGlasses outperforms existing baselines and variants among these tasks, highlighting the importance of active vision, as well as visual input and action representation choice.


\bibliography{ref}

\end{document}